\documentclass[11pt]{article}

\usepackage[letterpaper,margin=1in]{geometry}
\usepackage[T1]{fontenc}
\usepackage[utf8]{inputenc}
\usepackage{times}
\usepackage{amsmath,amssymb}
\usepackage{graphicx}
\usepackage{caption}
\usepackage{xcolor}
\usepackage{booktabs}
\usepackage{tabularx}
\usepackage{microtype}
\usepackage{inconsolata}
\usepackage{tcolorbox}
\tcbuselibrary{breakable,skins,listings}
\usepackage{iftex}

\ifPDFTeX\else
  \makeatletter
  \lst@AddToHook{Init}{%
    \catcode"2014=\active
    \catcode"2260=\active
    \catcode"2192=\active
    \catcode"03C3=\active
  }
  \makeatother
\fi

\tcbset{chemtext/.style={%
  enhanced,
  breakable,
  listing only,
  fonttitle=\small\bfseries,
  coltitle=black,
  colback=black!2,
  colbacktitle=black!8,
  colframe=black!35,
  boxrule=0.5pt,
  arc=2pt,
  left=10pt,right=10pt,top=8pt,bottom=8pt,
  before skip=12pt,after skip=12pt,
  listing options={%
    basicstyle=\ttfamily\fontsize{9}{10.5}\selectfont,
    aboveskip=0pt,belowskip=0pt,
    columns=fullflexible,
    keepspaces=true,
    breaklines=true,
    breakatwhitespace=true,
    breakindent=0pt,
    showstringspaces=false,
    upquote=true,
    tabsize=2,
    literate={×}{{\ensuremath{\times}}}1
             {≠}{{\ensuremath{\ne}}}1
             {→}{{\ensuremath{\rightarrow}}}1
             {—}{{\textemdash}}1
             {σ}{{\ensuremath{\sigma}}}1
  }
}}

\newtcbinputlisting[auto counter,number within=section]{\promptinput}[3][]{%
  chemtext,
  listing file={#3},
  title={Prompt~\thetcbcounter: #2},
  title after break={Prompt~\thetcbcounter: #2 (continued)},
  #1
}

\newtcbinputlisting[auto counter,number within=section]{\exampleinput}[3][]{%
  chemtext,
  listing file={#3},
  title={Example~\thetcbcounter: #2},
  title after break={Example~\thetcbcounter: #2 (continued)},
  #1
}

\usepackage{natbib}
\definecolor{linkgray}{rgb}{0.12,0.12,0.12}
\usepackage[colorlinks=true,linkcolor=linkgray,citecolor=linkgray,urlcolor=linkgray]{hyperref}

\makeatletter
\def\@fnsymbol#1{\ensuremath{\ifcase#1\or \#\or *\or \dagger\or \ddagger\or \S\or \P\or \|\or **\or \dagger\dagger\else\@ctrerr\fi}}
\makeatother

\title{ChemInsight: A Comprehensive Benchmark for Evaluating and Diagnosing
Chemistry Understanding and Reasoning in Large Language Models}

\author{
  Changyu Hu\textsuperscript{1,2}\thanks{These authors contributed equally.},\quad
  Yiming Zou\textsuperscript{1,2}\footnotemark[1],\quad
  Xin Tang\textsuperscript{1,3}\footnotemark[1],\quad
  Yiru Pan\textsuperscript{4},\quad
  Zicheng Huang\textsuperscript{5},\\[0.2em]
  Deng Zhao\textsuperscript{1},\quad
  Qing Cui\textsuperscript{1},\quad
  Zhenduo Zhang\textsuperscript{1}\thanks{Corresponding authors. Email: \href{mailto:zhenduo.zzd@antgroup.com}{\texttt{zhenduo.zzd@antgroup.com}}, \href{mailto:renwang.rw@antgroup.com}{\texttt{renwang.rw@antgroup.com}}, \href{mailto:xiaoli@pku.edu.cn}{\texttt{xiaoli@pku.edu.cn}}.},\quad
  Wang Ren\textsuperscript{1}\footnotemark[2],\quad
  Xiao Li\textsuperscript{2}\footnotemark[2] \\[0.4em]
  \small \textsuperscript{1}Ant Group \\[0.1em]
  \small \textsuperscript{2}Peking University \\[0.1em]
  \small \textsuperscript{3}University of Electronic Science and Technology of China \\[0.1em]
  \small \textsuperscript{4}Westlake University \\[0.1em]
  \small \textsuperscript{5}Peking University Health Science Center
}
\date{}

\begin{document}
\maketitle

\begin{abstract}
We introduce ChemInsight, a benchmark designed to evaluate large language models (LLMs) on chemistry tasks that require multi-step reasoning and structured chemical understanding. We propose a two-dimensional scoring system---scope and depth---to assess both task coverage and reasoning completeness. ChemInsight consists of 203 challenging chemistry questions organized into 50 bundles, which support the two-dimensional scoring evaluation scheme.

We evaluate ten representative LLMs using ChemInsight and observe consistently low performance across models, with scope scores below 30\% and depth scores below 11\%, substantially lower than human expert performance. We then proceed to ask (1) What factor limits the advancement of current LLMs in chemistry? (2) To what extent are statistically driven language-model training methods effective at modeling chemistry symbolic language such as SMILES? And (3) Can LLM performance in chemistry be improved through intensive training on large, domain-specific corpora?
\end{abstract}

\section{Introduction}

Benchmarks for large language models (LLMs) are designed to measure a model's capability within a specific domain. Recent advances in LLMs have led to remarkable success in tasks such as code generation~\citep{Surveyonllmcoding} and general question answering~\citep{Surveyonllqa}, substantially enhancing productivity across many industries~\citep{GenAIinhighereducation}. However, comparisons across standard tests and benchmark scores reveal that current LLMs are considerably weaker in chemistry than in domains such as history or programming, and they continue to underperform relative to human experts on chemistry tasks~\cite{OpenAI2025GPT5,zhao2025superchem}. This gap limits the reliability of LLMs for integration into rigorous chemistry research workflows and authentic educational settings.

Existing chemistry benchmarks provide only limited insight into model performance, typically reporting a single numerical score~\cite{rein2024gpqa,mirza2025framework,zhao2025superchem}. Such a score obscures the underlying bottlenecks that constrain LLM performance in chemistry and offers little guidance for targeted model improvement. Before LLMs can be responsibly adopted or systematically enhanced for chemistry applications, a more fine-grained and diagnostic evaluation of their capabilities is required.

\begin{quote}
\emph{``Measurement is the first step that leads to control and eventually to improvement.''} \\
--- H.\ James Harrington
\end{quote}

\begin{figure}[t]
  \centering
  \includegraphics[width=\textwidth]{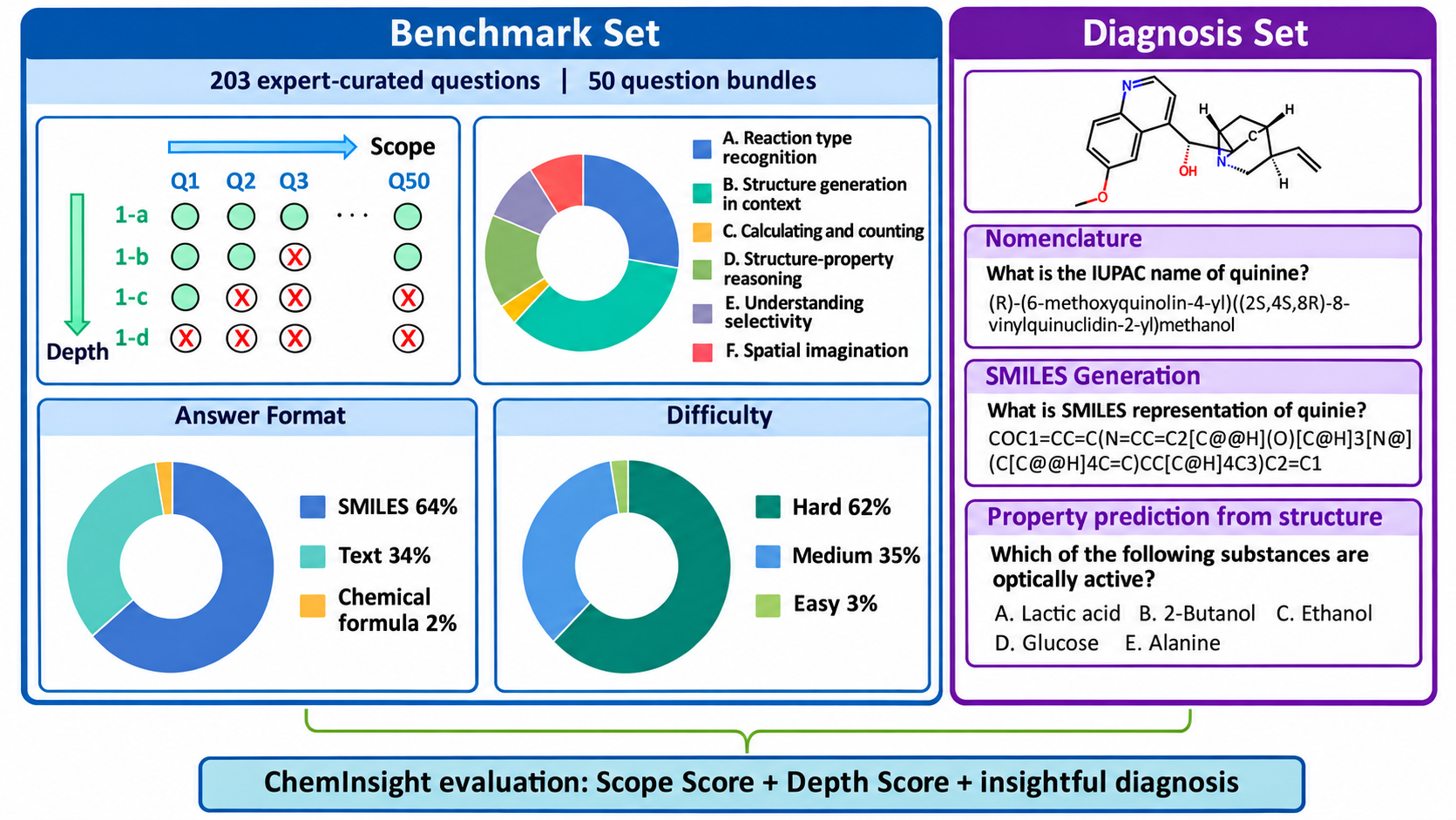}
  \caption{Overview of the ChemInsight evaluation framework.
    The Benchmark Set (left) contains 203 questions in 50 bundles covering six chemistry ability dimensions; the lower panels summarize answer formats and difficulty levels.
    The scope score measures overall performance, while the depth score rewards completion of interconnected questions within essence bundles.
    The Diagnosis Set (right) contains 50 single-step questions on IUPAC naming, SMILES generation, and property prediction to probe foundational chemistry skills.}
  \label{fig:experiments}
\end{figure}

In this work, we propose \textbf{ChemInsight}, a benchmark designed to support two-dimensional scoring and diagnostic analysis. ChemInsight consists of two components: a \emph{benchmark set} and a \emph{diagnosis set}. The benchmark set contains 203 challenging chemistry questions organized into 50 question bundles. Within each bundle, questions are interrelated and progressively increase in difficulty, resembling problems encountered in authentic chemistry research or advanced coursework. Successfully completing an entire bundle is treated as evidence of deep understanding and high task competency, and is therefore rewarded with an additional depth score. Model performance on the benchmark set is quantified using two dimensions: a \emph{scope score}, defined as the total proportion of correctly answered questions, and a \emph{depth score}, which captures how deeply a model can progress through a complex, multi-step chemistry task.

The diagnosis set comprises 50 single-step questions targeting three fundamental chemistry skills: nomenclature, SMILES generation, and property prediction. Each question isolates a single elementary skill. Unlike traditional benchmarks, this diagnostic component is explicitly designed to facilitate interpretation of model performance on the more complex, multi-skill questions in the benchmark set, enabling finer-grained analysis of model strengths and weaknesses.

We evaluate ten state-of-the-art LLMs using ChemInsight. The resulting scope scores range from 11.6\% to 27.8\%, while depth scores range from 2.3\% to 10.7\% (Figure~\ref{fig:Scope_depth}). Gemini 3 Pro achieves the highest performance, with a scope score of 27.8\% and a depth score of 10.7\%. Overall, the consistently low scores in both dimensions indicate that current LLM performance in chemistry remains unsatisfactory, particularly given that eight of the evaluated models represent the latest reasoning models. We further analyze model behavior on the diagnosis set and conduct a detailed examination of model outputs. In addition to reporting scores, the goal of this work is to provide an insightful diagnostic assessment of LLM capabilities in the chemistry domain. By identifying the key factors that limit LLM performance, we aim to illuminate pathways for future improvements in LLMs applied to scientific domains.

\section{Methodology}

\subsection{ChemInsight Benchmark Question Construction}

To evaluate the chemistry-specific capabilities of large language models (LLMs) and to provide diagnostic insights into their strengths and limitations, we construct two complementary question sets: a \textit{Benchmark Set} and an \textit{Diagnosis Set}.

\paragraph{Benchmark Set.}
The Benchmark Set consists of 203 original, challenging, and search-engine--resistant questions designed to require multi-step reasoning. These questions are organized into 50 bundles, with a primary focus on organic chemistry. Each question may draw upon multiple chemical principles and demand complex reasoning and/or quantitative calculations, thereby reflecting the types of problem-solving tasks encountered in advanced chemistry coursework and authentic research settings.

The design of the Benchmark Set follows two key standards:
(1) the complexity and difficulty are intended to be challenging even for domain experts, which in this work refers to chemistry majors at top-ranked universities; and
(2) the answers cannot be directly retrieved from general or professional chemistry databases or search engines, such as SciFinder or Reaxys.

The questions mainly test the following six major categories of chemical ability: A. reaction-type recognition, B. structure generation in context, C. calculation and counting, D. structure-property reasoning, E. selectivity prediction, and F. spatial imagination and three-dimensional thinking.

\paragraph{Diagnosis Set.}
The Diagnosis Set comprises 50 single-step questions that are intentionally simpler. These questions are designed to isolate and assess individual elementary chemistry skills, thereby facilitating the interpretation of model performance on the more complex, multi-skill questions in the Benchmark Set. Specifically, 30 questions explicitly test structure recognition and 20 questions assess structure-property reasoning.

All questions in both sets require LLMs to generate free-form answers; no multiple-choice questions are included. Both the Benchmark Set and the Diagnosis Set were independently created and cross-validated by domain experts to ensure correctness, clarity, and appropriate difficulty.

\subsection{Evaluation Method}

\paragraph{Baseline Models.}
\begingroup
\setlength{\emergencystretch}{3em}
In this benchmark, we evaluate 10 representative Large Language Models (LLMs) that cover current industry standards. Eight are evaluated in thinking mode:
GPT-5~\citep{OpenAI2025GPT5},
Gemini 3 Pro~\citep{Gemini2025},
Gemini 2.5 Pro~\citep{comanici2025gemini25pushingfrontier},
Ring-1T~\citep{lingteam2025stepevolvesscalingreinforcement},
Qwen3-235B-A22B-Thinking-2507~\citep{Qwen2025Thinking2507},
DeepSeek-V3.2~\citep{deepseekai2025deepseekv32pushingfrontieropen},
DeepSeek-V3.2-Speciale~\citep{deepseekai2025deepseekv32pushingfrontieropen},
and DeepSeek-R1~\citep{deepseekai2025deepseekr1incentivizingreasoningcapability}.
Additionally, we evaluate Ling-1T~\citep{lingteam2025activationboostedscalinggeneral} and DeepSeek-V3.1~\citep{DeepSeek2025V31} in non-thinking mode.
\par\endgroup

\paragraph{Response Generation Protocol.}
To ensure standardized assessment within complex problem scenarios, we designed a \textit{Task-Specific Structured Prompt} to guide the baseline models (the full prompt is provided in Appendix~\ref{sec:appendix2}). To mitigate generation randomness and assess model stability, we implement an \textbf{8-shot sampling strategy}, independently sampling 8 responses for each problem from every model.

\paragraph{Hybrid Evaluation Pipeline.}
Upon collecting the generated responses, we employ a hybrid evaluation mechanism tailored to different question types:

\textbf{1) Rule-Based Validation.}
For questions requiring molecular structure outputs (SMILES representations), we utilize the RDKit library. Due to the non-uniqueness of SMILES syntax, all generated SMILES strings are first canonicalized. We then verify exact matches against the canonicalized reference answers to determine correctness.

\textbf{2) LLM-as-a-Judge.}
To achieve high-quality, fine-grained evaluation for short-answer questions involving multi-step reasoning, we adopt the \textit{LLM-as-a-Judge} paradigm. We utilize a strong large language model as an evaluator, guided by an \textit{Evaluation-Specific Structured Prompt} (the full prompt is provided in Appendix~\ref{sec:appendix2}). To standardize scoring and reduce bias, the judge model follows a hierarchical evaluation process:
\begin{itemize}
    \item \textbf{Special Rule Enforcement \& Keyword Matching:} This phase serves as the primary filter. The judge first inspects the \textit{Note} field to enforce hard constraints, where a violation of special instructions (e.g., incorrect number of answers) triggers an immediate zero score regardless of content quality. Subsequently, it performs keyword matching, assessing the response based on the semantic coverage of essential terms defined in the reference.

    \item \textbf{Three-Level Scoring Rubric:} In general scenarios, the judge applies a rigorous tiered rubric: (1) \textbf{Full Credit} is awarded for responses that are conceptually complete, scientifically accurate, and logically consistent with the reference; (2) \textbf{Partial Credit} reflects constrained partial understanding. Here, we enforce a strict \textit{Anti-Hallucination Constraint}: partial scores are valid \textit{only} if the response remains strictly relevant without introducing extraneous concepts not mentioned in the question or reference; (3) \textbf{Zero Credit} is assigned for factual errors, violations of chemical principles, format non-compliance, or the inclusion of irrelevant information.
\end{itemize}

\paragraph{Score Aggregation.}
The final score $S$ for a model on a specific sub-question is calculated as the average of the scores obtained from the 8 sampling attempts:
\begin{equation}
    S = \frac{1}{K} \sum_{k=1}^{K} E(a_k, a_{\mathrm{ref}})
\end{equation}
where $K=8$, $a_k$ is the $k$-th generated answer, $a_{\mathrm{ref}}$ is the reference answer, and $E$ denotes the evaluation function (either Rule-based or LLM-based depending on the question type).

\subsection{Scoring System}
We propose a two-dimensional evaluation framework consisting of a \textit{Scope Score} and a \textit{Depth Score}, designed to jointly characterize the breadth and depth of chemical understanding demonstrated by large language models (LLMs).

The \textbf{Scope Score} ($S_{\mathrm{scope}}$) is defined as the raw accuracy percentage achieved by an LLM on the Benchmark Set. The benchmark contains a total of 203 questions, worth 262 points in total. The Scope Score is normalized to a maximum of 100. The more questions answered correctly, the higher the score that a model receives.

The \textbf{Depth Score} ($S_{\mathrm{depth}}$) captures the extent to which a model demonstrates coherent, multi-step understanding within problem groups. Unlike the scope score, the questions do not contribute equally to $S_{\mathrm{depth}}$. Instead, models earn a higher depth score by correctly answering multiple questions within the same \textit{essence bundle}. Among the 50 question bundles in the benchmark, 23 are designated as \textit{essence bundles}, in which questions are intentionally interconnected and arranged in increasing order of difficulty, thereby mimicking the progressive reasoning challenges encountered by human chemists in real-world scientific scenarios.

The depth score for an individual essence bundle $Q_i$ is defined as:
\begin{equation}
S_{\mathrm{depth}}(Q_i) = 2\,S_{\mathrm{scope}}(Q_i)\,\sin^3\!\left(\frac{\pi \omega_i}{2}\right),
\end{equation}
where $S_{\mathrm{scope}}(Q_i)$ denotes the Breadth Score contribution of the bundle $Q_i$, and $\omega_i$ represents the completion rate of that bundle. Full completion of an essence bundle ($\omega_i = 1$) indicates high competence in the corresponding chemical domain and yields the maximum depth score, which is twice the scope score contribution of that bundle.

The total Depth Score is computed as:
\begin{equation}
S_{\mathrm{depth}} = \sum_i S_{\mathrm{depth}}(Q_i).
\end{equation}

Because the Depth Score is designed to reward concentrated, multi-step mastery beyond isolated correctness, the combined score $S_{\mathrm{scope}} + S_{\mathrm{depth}}$ is not capped at 100.

\section{Experiments}

We investigate the following research questions (RQs):

\noindent\textbf{RQ1:} How do representative LLMs perform on ChemInsight in terms of both scope and depth? \\
\textbf{RQ2:} What limitations currently hinder the advancement of LLMs in the chemistry domain? \\
\textbf{RQ3:} To what extent are statistical language models effective at modeling chemistry symbolic language such as SMILES?

\begin{table}[t]
  \centering
  \footnotesize
  \setlength{\tabcolsep}{4pt}
  \renewcommand{\arraystretch}{1.05}
  \begin{tabular*}{\textwidth}{@{\extracolsep{\fill}}l c *{8}{c}@{}}
    \toprule
    \textbf{Model} & \shortstack{\textbf{Inference}\\\textbf{Mode}} & \shortstack{\textbf{Scope}\\\textbf{Score}} & \shortstack{\textbf{Depth}\\\textbf{Score}} & \textbf{A} & \textbf{B} & \textbf{C} & \textbf{D} & \textbf{E} & \textbf{F} \\
    \midrule
    \mbox{Gemini 3 Pro} & Thinking & 0.278 & 0.107 & 0.217 & 0.202 & 0.548 & 0.396 & 0.291 & 0.191 \\
    \mbox{GPT-5} & Thinking & 0.215 & 0.069 & 0.076 & 0.078 & 0.275 & 0.450 & 0.360 & 0.207 \\
    \mbox{Gemini 2.5 Pro} & Thinking & 0.212 & 0.059 & 0.100 & 0.106 & 0.371 & 0.421 & 0.304 & 0.181 \\
    \mbox{DeepSeek-V3.2-Speciale} & Thinking & 0.208 & 0.067 & 0.120 & 0.120 & 0.370 & 0.354 & 0.279 & 0.163 \\
    \mbox{DeepSeek-V3.2} & Thinking & 0.181 & 0.047 & 0.089 & 0.091 & 0.359 & 0.351 & 0.244 & 0.156 \\
    \mbox{Qwen3-235B-A22B-Thinking-2507} & Thinking & 0.151 & 0.034 & 0.058 & 0.063 & 0.117 & 0.325 & 0.262 & 0.127 \\
    \mbox{Ring-1T} & Thinking & 0.151 & 0.028 & 0.052 & 0.051 & 0.257 & 0.318 & 0.275 & 0.137 \\
    \mbox{DeepSeek-R1} & Thinking & 0.144 & 0.021 & 0.056 & 0.061 & 0.225 & 0.292 & 0.225 & 0.130 \\
    \mbox{Ling-1T} & Non-thinking & 0.122 & 0.021 & 0.009 & 0.003 & 0.295 & 0.320 & 0.124 & 0.155 \\
    \mbox{DeepSeek-V3.1} & Non-thinking & 0.116 & 0.024 & 0.015 & 0.007 & 0.238 & 0.300 & 0.130 & 0.160 \\
    \bottomrule
  \end{tabular*}
  \caption{\label{tab:model_performance}
    Performance comparison of different models. Inference Mode indicates the mode used in this evaluation.
    Scope and Depth scores are normalized by 262 and 524 respectively.
    Columns A--F are normalized by 137, 172, 17, 78, 46, and 48.5 respectively.}
\end{table}

\subsection{Model Performance on ChemInsight (RQ1)}

\begin{figure}[t]
  \centering
  \begin{minipage}[t]{0.485\textwidth}
    \vspace{0pt}
    \centering
    \includegraphics[width=\linewidth]{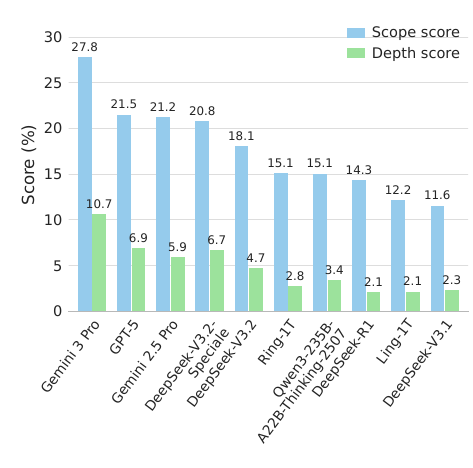}
    \caption{Scope and depth scores on the ChemInsight Benchmark Set. Scores are averaged over eight responses per question and normalized by 262 and 524 points, respectively.}
    \label{fig:Scope_depth}
  \end{minipage}\hfill
  \begin{minipage}[t]{0.485\textwidth}
    \vspace{0pt}
    \centering
    \includegraphics[width=\linewidth]{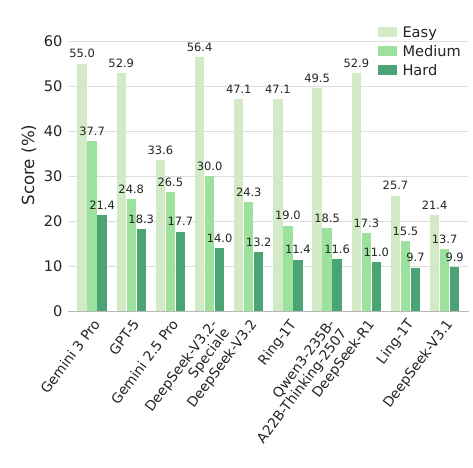}
    \caption{Benchmark performance by expert-assigned question difficulty. Scores generally decrease from easy to medium and hard questions.}
    \label{fig:difficulty}
  \end{minipage}
\end{figure}

The performance of the evaluated LLMs on the ChemInsight benchmark, averaged over eight independent runs, is summarized in Table~\ref{tab:model_performance} and visualized in Figure~\ref{fig:Scope_depth}. Overall, the results indicate that all evaluated models achieve relatively low scores on this newly proposed benchmark. Scope scores range from 11.6\% to 27.8\%, with none exceeding 50\%. Depth scores are consistently low as well, ranging from 2.3\% to 10.7\% in the ten models.

Among all evaluated models, \textit{Gemini 3 Pro} attains the highest performance in both scope and depth, with scores of 27.8\% and 10.7\%, respectively. Nevertheless, this performance remains substantially below that of human experts, who achieve a 36.3\% scope score. Reasoning-oriented models generally outperform non-reasoning models, as evidenced by \textit{Ling-1T} and \textit{DeepSeek-V3.1}, which rank near the bottom in both scope and depth metrics. It is also observed from Figure~\ref{fig:difficulty} that scores decrease with increasing human-expert-defined difficulty level.

The depth score introduces an additional dimension to evaluate and differentiate model performance. For example, \textit{GPT-5} and \textit{Gemini 2.5 Pro} obtain nearly identical scope scores (21.5\% and 21.2\%, respectively). Based solely on scope, one might conclude that the two models exhibit comparable capabilities. However, the depth score reveals a meaningful distinction: GPT-5 achieves a higher depth score (6.9\%) than Gemini 2.5 Pro (5.9\%), suggesting that GPT-5 is able to engage more deeply with complex chemistry problems.

In this evaluation work, depth scores appear to be roughly proportional to scope scores and exhibit only limited variation. This behavior is primarily attributable to the overall poor performance of existing LLMs on ChemInsight. As artificial intelligence technology continues to advance, scope scores are expected to increase and eventually approach saturation. At that stage, the depth score will become a more informative and discriminative metric for distinguishing among high-performing models.

\subsection{Shortcomings Limiting LLM Advancement in Chemistry (RQ2)}

Despite their strong performance in domains such as programming and natural language processing, large language models (LLMs) exhibit markedly weaker capabilities in chemistry. This raises a fundamental question: \emph{what factors limit the effectiveness of LLMs in the chemical domain?}

\begin{figure}[t]
  \centering
  \begin{minipage}[t]{0.485\textwidth}
    \vspace{0pt}
    \centering
    \includegraphics[width=\linewidth]{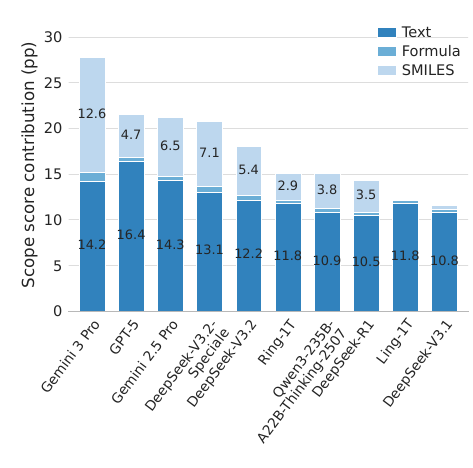}
    \caption{Contributions of text, chemical-formula, and SMILES answers to the total scope score. Segment heights are percentage points of the normalized benchmark score.}
    \label{fig:format}
  \end{minipage}\hfill
  \begin{minipage}[t]{0.485\textwidth}
    \vspace{0pt}
    \centering
    \includegraphics[width=\linewidth]{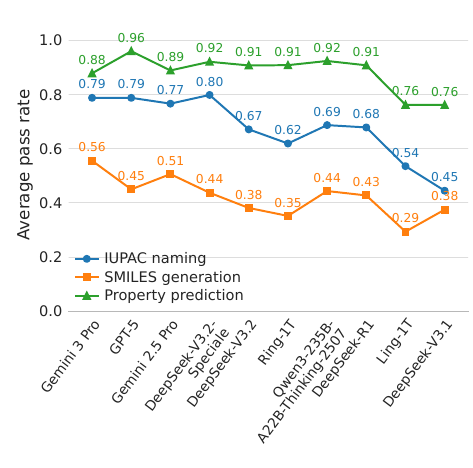}
    \caption{Accuracy on the Diagnosis Set for IUPAC naming, SMILES generation, and property prediction. Performance is strongest on property prediction and weakest on SMILES generation.}
    \label{fig:diagnosis}
  \end{minipage}
\end{figure}

Our evaluation results indicate that \textbf{in-context comprehension and generation of SMILES representations constitute the primary bottleneck}. The required answer formats in the ChemInsight benchmark fall into three categories: natural language text, chemical formulas, and structural representations expressed as SMILES. Performance across these categories is summarized in Figure~\ref{fig:format}. While most models achieve comparable scores on text-based and chemical-formula questions, substantial performance differences emerge on SMILES-format questions, which clearly distinguish among models.

Variation in SMILES performance accounts for approximately 82\% of the variance in the overall benchmark score, highlighting its dominant influence on total model performance.

\begin{figure}[t]
  \centering
  \includegraphics[width=0.80\textwidth]{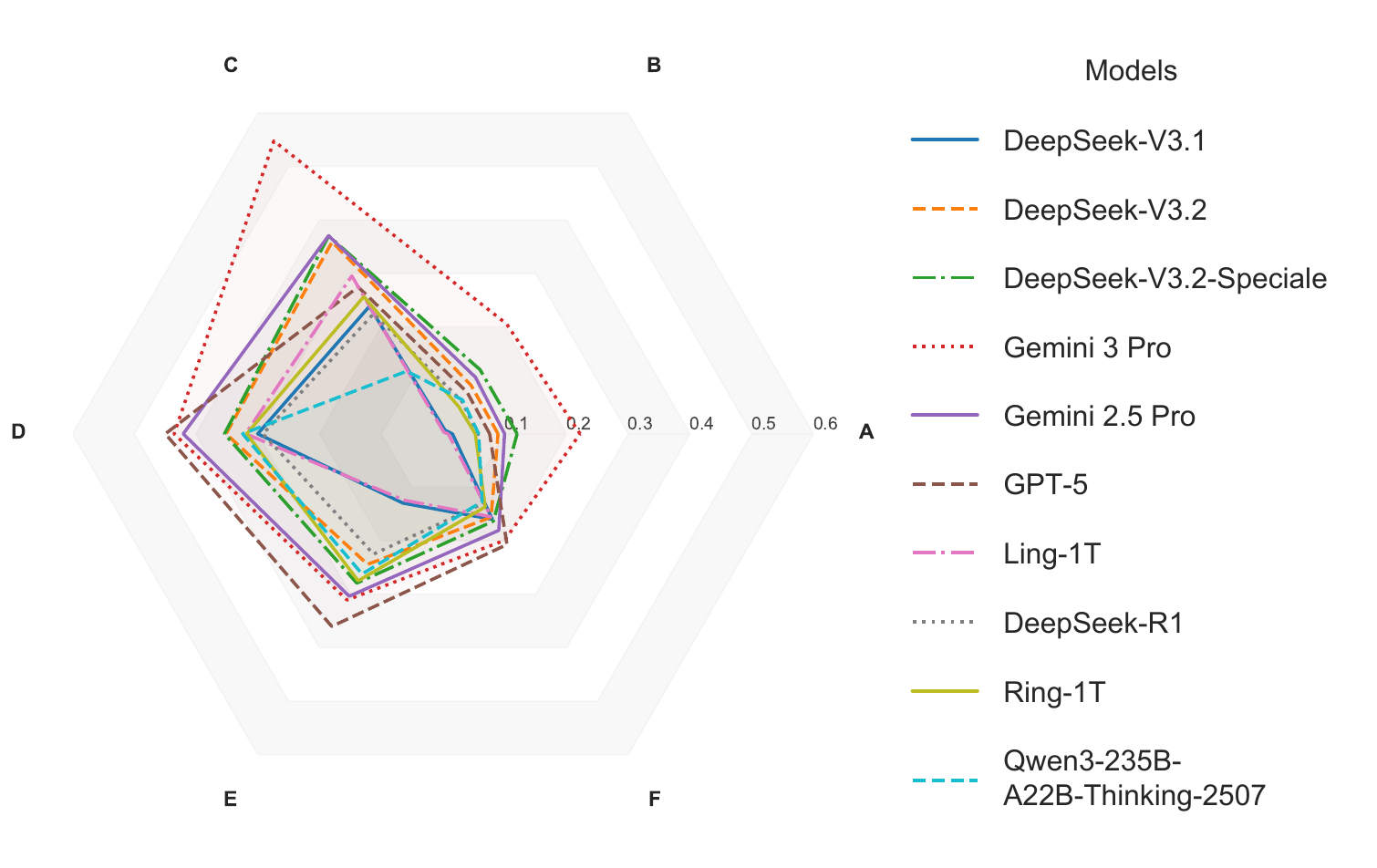}
  \caption{Normalized performance across the six chemistry ability dimensions in ChemInsight: (A) reaction-type recognition, (B) structure generation in context, (C) calculation and counting, (D) structure-property reasoning, (E) selectivity prediction, and (F) spatial imagination and three-dimensional thinking. Each curve represents one model; higher values indicate stronger performance.}
  \label{fig:radar}
\end{figure}

Although the benchmark questions are designed to be comprehensive, each question is annotated by its primary objective and grouped into six core chemical ability dimensions: (A) reaction-type recognition, (B) structure generation in context, (C) calculation and counting, (D) structure-property reasoning, (E) selectivity prediction, and (F) spatial imagination and three-dimensional thinking. The radar plot shown in Figure~\ref{fig:radar} reveals pronounced weaknesses across nearly all dimensions, with the notable exception of calculation and counting. Importantly, this is the only ability that does not rely heavily on chemical structure comprehension.

In the ChemInsight benchmark, chemical structures are predominantly represented using SMILES notation. The observed pattern of results therefore provides strong evidence that deficient SMILES comprehension and manipulation is a central limitation of current LLMs, and a key factor constraining their progress in chemistry-focused reasoning tasks.

\subsection{Chemistry SMILES Representation Competency Diagnosis (RQ3)}

Chemistry-domain corpora typically contain a mixture of natural language text and symbolic chemical representations.

In the Benchmark Set, questions involving SMILES are often coupled with additional requirements such as contextual understanding or multi-step reasoning. As a result, poor model performance on these questions cannot be confidently attributed to a single underlying competency. To address this limitation, we construct a Diagnosis Set designed to evaluate isolated chemistry skills in a controlled manner.

Importantly, the answers to these single-skill questions can be directly retrieved from authoritative chemistry databases. Human experts, state-of-the-art search engines, and professional software tools (e.g., ChemDraw) can achieve near-perfect accuracy on these tasks. While such questions do not resemble real-world chemistry problem-solving scenarios, they are intentionally designed to diagnose specific foundational competencies of LLMs in chemical representation and reasoning.

The evaluation results (Figure~\ref{fig:diagnosis}) indicate that most assessed LLMs demonstrate strong competency in property prediction. Reasoning-oriented models achieve approximately 90\% accuracy, while non-reasoning models reach around 76\%.

These results suggest that once a chemical compound is recognized, LLMs are generally capable of associating it with relevant chemical properties---an important criterion for structural understanding.

In chemistry, a compound may be represented in multiple forms, including its common name, IUPAC systematic name, structural diagram, and SMILES string. For example, the compound \emph{quinine} can be referred to by its common name or by its IUPAC name, \emph{(R)-(6-methoxyquinolin-4-yl)((2S,4S,8R)-8-vinylquinuclidin-2-yl)methanol}. It can also be represented as a structural formula illustrating atomic connectivity and spatial arrangement, or as a SMILES string used by professional software to render molecular structures.

The results show that model accuracy ranges from 44.5\% to 78.8\% when translating common names to IUPAC names, and drops further to 29.4\%--56.6\% when translating common names to SMILES. These findings indicate that competency in chemical symbolic language---particularly SMILES representation---remains a major technical bottleneck. This deficiency substantially limits the ability of LLMs to perform complex and multi-step chemistry tasks that require precise manipulation of molecular structures.

\section{Related Work}

\textbf{Benchmarks on chemistry.} Currently, there exists a significant body of work regarding chemistry benchmarks. \textbf{GPQA}~\citep{rein2024gpqa} encompasses the three major fields of biology, physics, and chemistry, comprising 448 high-quality multiple-choice questions written and verified by PhD-level experts in these fields. It challenges the knowledge boundaries of large models in cutting-edge scientific domains. \textbf{ChemBench}~\citep{guo2023can} establishes a comprehensive benchmark suite containing eight chemistry tasks, categorizing the capabilities of LLMs in chemistry into three core domains: understanding, reasoning, and explanation. This provides a framework for evaluating the ``logical depth'' of LLMs in the scientific domain. GPT-4 achieved 71\% and Claude-3.5 achieved 61\% on ChemBench as reported. \textbf{SUPERChem}~\citep{zhao2025superchem} provides expert-authored standardized solution paths for each question, employing the Reasoning Path Fidelity (RPF) metric to evaluate whether the model's logical process in solving the problem is correct. By combining RPF scores with accuracy, it enables an accurate assessment of the comprehensive reasoning capabilities of the models.

\textbf{LLM in chemistry.} With the continuous advancement of AI technology, the application of AI in chemistry has been extensively explored and developed. LLMs play a dual role in chemical applications~\citep{ramos2025review}: acting as chemical tools and functioning as drivers for autonomous agents. \textbf{ChemLLM}~\citep{zhang2024chemllm} is a conversational large language model specifically designed for the field of chemistry, aiming to address the performance deficiencies of general-purpose large models when processing specialized chemical knowledge. Through the construction of chemistry-specific datasets and a comprehensive training pipeline, the research team achieved the integration of knowledge across all chemical disciplines within the large language model framework, endowing the model with the ``chemical intuition'' required to handle complex interdisciplinary problems.

\section{Conclusion}

In this work, we introduced ChemInsight, a rigorous benchmark designed to evaluate the depth and scope of Large Language Models (LLMs) in the domain of chemistry. Through extensive experiments on representative models, including Gemini 3 Pro and GPT-5, we quantified the significant gap that remains between current state-of-the-art LLMs and human experts. Our findings reveal that while models demonstrate reasonable proficiency in general chemical textual knowledge, they struggle significantly with complex multi-step reasoning and deep conceptual understanding, as evidenced by the consistently low Depth Scores.

Crucially, our analysis identifies the comprehension and generation of symbolic representations---specifically SMILES---as the primary bottleneck restricting model performance. The symbolic nature of chemical language demands a precision that current general-purpose LLMs lack. \textbf{While the systematic training or fine-tuning of models on SMILES grammars and large-scale chemical corpora falls outside the scope of this benchmarking study, it represents a critical direction for future research.} Addressing this limitation may require novel tokenization strategies or neuro-symbolic approaches to bridge the gap between natural language processing and chemical structure manipulation.

Furthermore, we acknowledge that chemistry is an inherently multi-modal discipline, relying heavily on molecular graphs, spectra, and reaction diagrams to convey information. \textbf{Our current evaluation is predominantly based on language models processing text and serialized strings. However, a comprehensive understanding of chemistry necessitates the ability to perceive and interpret visual data.} Future work should extend the ChemInsight framework to include Multimodal Large Language Models (MLLMs), investigating whether visual encoders can enhance the reasoning capabilities required for tasks involving spatial imagination and structural analysis.

\section*{Limitations}

Despite the comprehensive nature of ChemInsight, several limitations should be noted. First, as highlighted in the conclusion, our benchmark currently focuses on the textual and symbolic modalities (SMILES/InChI). This excludes a vast array of chemical problems that require visual interpretation, such as reading NMR spectra or analyzing steric hindrance from 3D molecular representations. Consequently, the performance reported here may not fully reflect a model's potential utility in a laboratory setting where visual data is ubiquitous.

Second, although we employed a rigorous prompt structure and expert-verified answers, the evaluation of open-ended generation tasks by LLM-based judges inevitably carries a degree of uncertainty. While we mitigated this through structured rubrics and multiple evaluation runs, subtle correct reasoning paths that deviate from standard reference answers might still be under-penalized or over-penalized.

Finally, the field of LLMs is evolving rapidly. The ``Scope'' and ``Depth'' scores reported in this paper represent a snapshot of model capabilities at the time of writing. As models with stronger reasoning capabilities (e.g., ``System 2'' thinking modes) continue to emerge, the difficulty ceiling of the benchmark may need to be dynamically adjusted to maintain its discriminative power.

\section*{Acknowledgments}
This work was carried out by Changyu Hu, Yiming Zou and Xin Tang during their internship at Ant Group. 

\bibliography{custom}

\clearpage
\appendix

\section{Benchmark Instance Gallery}
\label{sec:appendix1}

The Benchmark Set consists of 203 original, challenging, and search-engine--resistant questions designed to require multi-step reasoning. To facilitate a deeper understanding of the ChemInsight benchmark, Example~\ref{box:ess_question} and Example~\ref{box:ordinary_question} present two question bundles from the benchmark set. These examples demonstrate tasks that integrate natural language descriptions, chemical logic, and symbolic representations. Each example includes prior knowledge, the input question, the verified ground truth, and scoring metadata.

\exampleinput[label=box:ess_question]
  {Essence Question Bundle}
  {examples/essence.txt}

\exampleinput[label=box:ordinary_question]
  {Ordinary Question Bundle}
  {examples/ordinary.txt}

\clearpage
\section{Prompt Design}
\label{sec:appendix2}

To standardize model outputs within complex chemical scenarios, we designed a \textit{Task-Specific Structured Prompt} (Prompt~\ref{box:task_prompt}). For assessment, to achieve high-quality, fine-grained evaluation for short-answer questions involving multi-step reasoning, we adopt the LLM-as-a-Judge paradigm: we employ an \textit{Evaluation-Specific Structured Prompt} (Prompt~\ref{box:evaluation_prompt}) that guides the judge model through a hierarchical protocol.

\promptinput[label=box:task_prompt]
  {Task-Specific Structured Prompt}
  {prompts/task.txt}

\promptinput[label=box:evaluation_prompt]
  {Evaluation-Specific Structured Prompt}
  {prompts/evaluation.txt}

\end{document}